\documentclass[times, review, 10pt]{elsarticle}

\biboptions{sort&compress}
\usepackage{amssymb}
\usepackage{amsmath}
\usepackage{algorithmic}
\usepackage{algorithm}
\usepackage{array}
\usepackage{caption} 
\usepackage[caption=false,font=normalsize,labelfont=sf,textfont=sf]{subfig}
\usepackage{textcomp}
\usepackage{stfloats}
\usepackage{url}
\usepackage{verbatim}
\usepackage{multirow}
\usepackage{graphicx}
\usepackage{makecell} 
\usepackage{threeparttable}
\usepackage{enumitem}
\usepackage{csquotes}
\usepackage[table]{xcolor}
\definecolor{darkgreen}{RGB}{0, 100, 0}
\definecolor{red}{RGB}{238 44 44}
\usepackage[colorlinks,
linkcolor=red,
anchorcolor=blue,
citecolor=green
]{hyperref}
\usepackage{cleveref}
\Crefname{figure}{Fig.}{figures}
\crefname{figure}{Fig.}{figures}

\journal{Pattern Recognition}

\begin{document}

\begin{frontmatter}

\title{LSVG: Language-Guided Scene Graphs with 2D-Assisted Multi-Modal Encoding for 3D Visual Grounding}

\author[1]{Feng Xiao}
\author[2]{Hongbin Xu}
\author[3]{Guocan Zhao}
\author[1]{Wenxiong Kang\corref{cor1}}
\ead{auwxkang@scut.edu.cn}   
\cortext[cor1]{Corresponding author.}

\affiliation[1]{organization={School of Automation Science and Engineering, South China University of Technology},
            city={Guangzhou},
            postcode={510006}, 
            country={China}}
\affiliation[3]{organization={School of Future Technology, South China University of Technology},
            city={Guangzhou},
            postcode={510006}, 
            country={China}}
\affiliation[2]{organization={ByteDance Seed},
            city={Beijing},
            postcode={100000}, 
            country={China}}

\begin{abstract}
3D visual grounding aims to localize the unique target described by natural languages in 3D scenes. 
The significant gap between 3D and language modalities makes it a notable challenge to distinguish multiple similar objects through the described spatial relationships. 
Current methods attempt to achieve cross-modal understanding in complex scenes via a target-centered learning mechanism, ignoring the modeling of referred objects. 
We propose a novel 3D visual grounding framework that constructs language-guided scene graphs with referred object discrimination to improve relational perception. 
The framework incorporates a dual-branch visual encoder that leverages pre-trained 2D semantics to enhance and supervise the multi-modal 3D encoding. Furthermore, we employ graph attention to promote relationship-oriented information fusion in cross-modal interaction. The learned object representations and scene graph structure enable effective alignment between 3D visual content and textual descriptions. Experimental results on popular benchmarks demonstrate our superior performance compared to state-of-the-art methods, especially in handling the challenges of multiple similar distractors.
\footnote{The code will be released at \href{https://github.com/onmyoji-xiao/LSVG}{{https://github.com/onmyoji-xiao/LSVG}}.}
\end{abstract}




\end{frontmatter}


\section{Introduction}
Visual grounding plays an important role in 3D scene understanding, applied in various fields including robot navigation, autonomous driving, and human-computer interaction \cite{liu2025survey}. Given the textual description and 3D scene, 3D visual grounding aims to localize the unique target corresponding to the description. The main challenge lies in differentiating similar candidate objects within the same scene \cite{achlioptas2020referit3d}. 
It necessitates effective alignment between referential relationships in natural languages and the spatial positions in real scenes.

\begin{figure}[!t]
   \centering
    \includegraphics[width=0.7\linewidth]{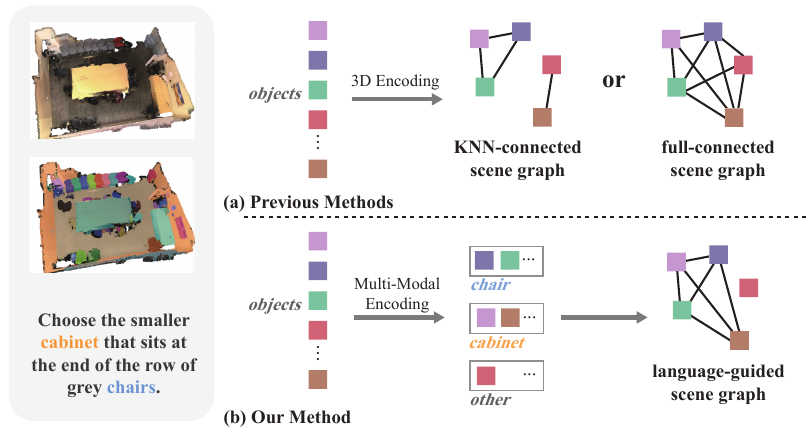}
    \caption{An example of text-guided localization in a scene. In the text, the target is marked in yellow and the referred object is marked in blue. Previous methods (a) construct a fully connected scene graph or use KNN to find objects with spatial neighbors for connection. Instead, our method (b) creates multi-modal representations for all objects and then constructs the scene graph by connecting objects that are identified as referred objects or targets through linguistic cues analysis.}
    \label{start}
    \vspace{-0.4cm}  
\end{figure}

Recent methods for 3D visual grounding attempt to independently encode the features of 3D point clouds and text into a shared feature space for 3D-language alignment \cite{liu2025survey}. 
He et al. \cite{he2021transrefer3d} first introduce the Transformer \cite{vaswani2017attention} architecture to establish entity-and-relation aware attention, then an increasing number of researchers are focusing on utilizing Transformer-based fusion to achieve the cross-modal interaction \cite{yang2021sat, roh2022languagerefer, huang2022multi, guo2023viewrefer, chen2022language}. In the latest methods, Xu et al. \cite{xu2024multi} and Chang et al. \cite{chang2024mikasa} respectively incorporate text-related attribute awareness and spatial awareness into the Transformer to achieve more accurate target localization. 
However, distinguishing targets from intra-class distractors remains challenging due to \textbf{insufficient relational reasoning} between 3D vision and language modalities. The scene graph provides a structured representation of the scene and allows reasoning by explicitly modeling object relationships and attributes \cite{chang2021comprehensive}. It plays a pivotal role in visual scene comprehension and has been extensively investigated in existing research \cite{zhang2026scenellm}. However, prior methods typically construct fully connected graphs \cite{feng2021free, chang2024mikasa, li2025r2g} or rely solely on KNN (K-Nearest Neighbors) for edge construction \cite{huang2021text, yuan2021instancerefer}. These designs lead to high computational costs due to too many nodes or fail to capture long-range relations. Treating all objects equally can also make it hard to distinguish similar relationships.

Understanding spatial relationships in 3D scenes can be divided into \textbf{two primary aspects}: (i) identifying objects with attributes such as color, shape, scale, and semantic categories corresponding to linguistic targets and referred objects; (ii) inferring the spatial structure between objects that satisfy the described relationships. 
The sparse, noisy, and incomplete characteristics of 3D point clouds lead to the lack of color information and texture details \cite{lei2023recent}, preventing them from fulfilling the object encoding in (i). Although many studies \cite{chen2020scanrefer, zhao20213dvg, luo20223d, cai20223djcg, chen2022d, zhang2023multi3drefer} encode 2D views for a more complete representation, they fail to bridge the semantic gap arising from the modality discrepancy between 3D visual and natural language representations. 
Instead of existing methods, our approach confines the scope of relational perception to the area between linguistically referred objects, as shown in \Cref{start} (b). 2D views are not only used to enhance point cloud representation but also to supervise 3D-language alignment. We aim to reinforce the crucial relations within the scene graph by creating multi-modal representations during the point cloud encoding process.

In this paper, we further explore scene graph construction in the 3D visual grounding task and propose a novel framework, \textbf{LSVG}, to address the challenge of multiple distractors. The primary innovations include a dual-branch 3D multi-modal encoder that requires only single-view guidance, and a language-guided scene graph on predicted referred objects. Rather than directly concatenating 2D projection features during point cloud encoding, we leverage the pre-trained 2D multi-modal model to enhance and supervise 3D visual encoding. The potential referred objects undergo a preliminary screening by aligning the class-aware prompts with object features derived from multi-modal encoding. The scene graphs with these objects as nodes only perform cross-modal information aggregation and relational reasoning for the relationship clues in the textual descriptions. The target node exhibits strong activation from its contextual connection with the correct referred objects and attains the highest prediction score. We evaluate our method on the widely used ReferIt3D \cite{achlioptas2020referit3d} and ScanRefer \cite{chen2020scanrefer} benchmarks, and it outperforms the existing state-of-the-art methods.

The main contributions are summarized as follows:
\begin{itemize}[leftmargin=*]
\item{We propose a new framework for 3D visual grounding via language-guided scene graphs. The framework aims to improve localization accuracy in complex scenes by enabling more precise identification of referred objects and relation-aware contextual reasoning with graph attention.}
\item{We design a dual-branch 3D visual encoder that employs a pre-trained 2D multi-modal model for feature enhancement and multi-modal supervision. This establishes a robust foundation for the effective construction of scene graphs.}
\item{Unlike prior works that rely on fully-connected or KNN-based graph structures, our scene graph edges are based on 3D-language semantic matching. This approach directly connects nodes that are potential referred objects or targets, reducing noise and computational overhead.}

\end{itemize}

\section{Related Work}
\subsection{3D Visual Grounding}
3D visual grounding, based on vision-language learning, is an important task in scene understanding, which has been extensively researched using two benchmarks: ScanRefer \cite{chen2020scanrefer} and ReferIt3D \cite{achlioptas2020referit3d}. Given a textual description and a 3D scene modeled by a point cloud or posed views, the multi-modal model can localize the described 3D object and produce a 3D bounding box. The prevailing computational frameworks for 3D visual grounding can be categorized into two primary architectural paradigms: (i) proposal-based two-stage methods that adopt a sequential paradigm of region proposal generation followed by fine-grained classification, and (ii) single-stage methods that implement end-to-end optimization through unified spatial prediction while concurrently addressing multi-modal feature alignment \cite{liu2025survey}. In this paper, we focus on investigating the 3D-language alignment problem of the second stage in two-stage localization. 

\subsubsection{Two-stage Localization}
The two-stage architecture typically employs dedicated instance segmentation or detection models to generate potential object proposals, such as VoteNet \cite{qi2019deep}, PointGroup \cite{jiang2020pointgroup}, and GroupFree \cite{liu2021group}. Existing methods propose distinct 3D visual perception modules and integrate textual features with visual features to identify optimal prediction targets. Recent studies reveal two main challenges: spatial description ambiguity due to variations in viewpoint, and the matching of point clouds and texts while dealing with interference from similar objects. LanguageRefer \cite{roh2022languagerefer} and VPP-Net \cite{shi2024aware} adapt viewpoint-dependent descriptions by predicting the orientation of observing viewpoints to adjust the position of point clouds. On the other hand, MVT \cite{huang2022multi} and ViewRefer \cite{guo2023viewrefer} encode multiple rotating scene information into visual features simultaneously to implicitly learn viewpoint consistency. Cross-modal feature alignment methods typically employ scene graphs to model text-to-object relevance. FFL-3DOG \cite{feng2021free} and TGNN \cite{huang2021text} measure text-node similarity through graph structures, while MiKASA \cite{chang2024mikasa} introduces multi-key-anchor fusion encoding for object-pair modeling. In contrast, R2G \cite{li2025r2g} constructs semantic concept-based graphs using heuristic rules.
Although the methods mentioned above achieve remarkable performance, they do not highlight the localization of referred objects, making relational learning in complex scenes still challenging.

\subsubsection{2D-3D Vision Fusion}
Recent advances in 3D visual grounding have demonstrated the benefits of incorporating 2D visual features to complement the sparse geometric information in point clouds. Semantic segmentation-based approaches \cite{chen2020scanrefer, zhao20213dvg, chen2022d, luo20223d} utilize lightweight architectures like ENet \cite{paszke2016enet} to establish dense 2D-3D correspondence by orthogonally projecting pixel-wise features onto point clouds. 3DJCG \cite{cai20223djcg} and MA2TransVG \cite{xu2024multi} project the multi-view features to object proposals that are subsequently fused with geometric primitives to form attribute representations. SAT \cite{yang2021sat} utilizes a pre-trained Faster-RCNN \cite{salvador2016faster} detector to extract the region of interest features from sampled frames for 2D semantically assisted alignment. LAR \cite{bakr2022look} synthesizes the 2D images from virtual cameras at five principal viewing directions of object proposals and extracts the features. Directly concatenating 2D features with 3D point clouds disrupts the original 2D-language alignment learned by vision-language models. Recent works like CADN \cite{ke2025language} highlight the importance of explicit cross-modal consistency in 2D visual grounding. Inspired by this, we extend such alignment to 3D scenes.

Our method derives the object features from just one view to enhance the 3D point cloud encoding and attain improved multi-modal alignment.

\subsection{Vision-Language Pretraining}
Pre-trained vision-language models learn joint visual and linguistic representations on large-scale data, widely used for downstream tasks related to multi-modal scene understanding \cite{chen2023vlp}. In the image-text domain, CLIP \cite{radford2021learning} is the most representative pre-trained model for cross-modal semantic alignment of image and text through contrast learning. ALIGN \cite{jia2021scaling} enhances generalization in long-tailed scenarios by scaling up contrastive learning through massive data augmentation. In contrast, ALBEF \cite{li2021align} employs a dual-stream architecture integrated with multi-task learning to strengthen cross-modal alignment through joint representation optimization. Furthermore, researchers also extend visual language pre-training to other application scenarios using contrastive learning as the core foundation \cite{chen2023clip2scene}.

In 3D multi-modal tasks, the scarcity of 3D data and costly annotation pose significant challenges for training on large-scale real-world datasets \cite{jia2024sceneverse}. Recent approaches leverage pre-trained 2D multi-modal models and conversion relations between 2D and 3D visual representations to enhance cross-modal perception of 3D vision and languages in 3D-input networks. ULIP \cite{xue2023ulip} uses frozen image and text encoders from CLIP to construct a unified representation of image, text, and 3D point cloud to improve the performance of 3D backbone networks. In this paper, we leverage the pre-trained knowledge of a 2D vision-language model to improve the alignment of point cloud features with text features, while incorporating multi-modal features as supplementary information to enhance the 3D visual representation.

\begin{figure*}[t]
    \centering{\includegraphics[width=1\linewidth]{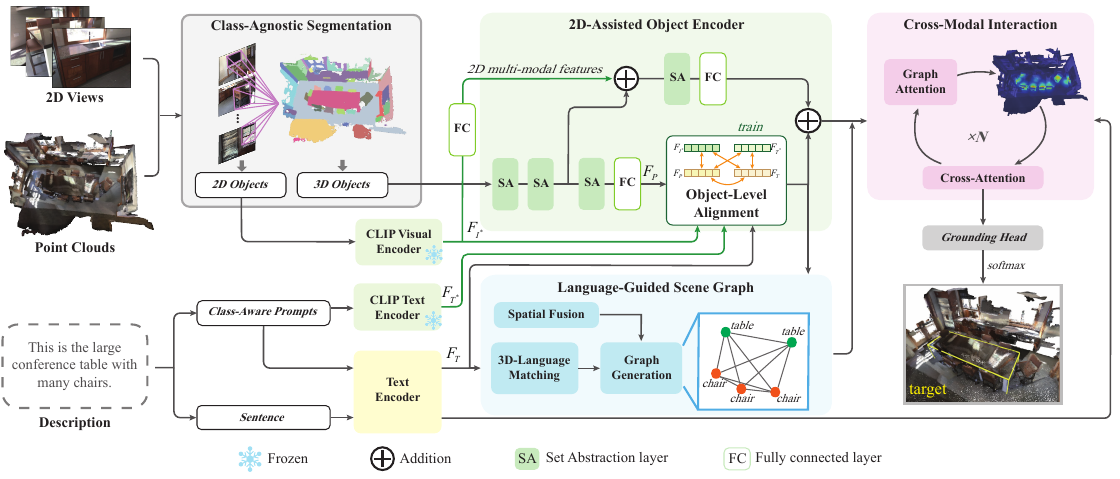}}
	\caption{\textbf{The overall architecture of our proposed LSVG.} The input scene point clouds are segmented into object point clouds, which are then processed by a dual-branch object encoder. We leverage 2D multi-modal features from the frozen CLIP model to guide additional feature encoding. Object-level feature alignment is calculated exclusively during training. Instance identifiers are separated from the original textual descriptions and expanded into class-aware semantic prompts. This enables precise 3D-language correspondences, which then create a scene graph that captures dependencies between objects. Cross-modal interaction is achieved through an iterative graph attention architecture. The grounding head implemented by the MLP outputs the target with the highest probability.}
   \vspace{-0.4cm}
    \label{fig1}
\end{figure*}

\section{Methods}

As shown in Figure \Cref{fig1}, our framework includes \textbf{a 2D-assisted dual-branch 3D object encoder, a language-guided scene graph, and a cross-modal interaction module}. The 3D point cloud network performs sampling and encoding for each object proposal in the scene. It utilizes a dual-branch architecture that supports 2D-supervised 3D-language alignment and feature enhancement. For benchmarks without pre-defined object bounding boxes, we employ the class-agnostic object segmentation model, MaskClustering \cite{yan2024maskclustering}, to attain instance segmentation free from semantic categories. The scene graph is constructed with the semantically relevant candidates as nodes, which are extracted through 3D-language matching. The cross-modal interaction module alternately performs object-text cross-attention and graph attention, ultimately predicting the target with the highest correlation to the description.

\subsection{Multi-Modal 3D Object Encoding}

Existing works mainly encode the point clouds of objects as pure visual features before performing cross-modal interactions. Due to insufficient 3D data and scenes, it is challenging for 3D vision and language information to achieve the same alignment performance as 2D-language learning trained on large-scale data. To address this problem, we utilize the guidance of aligned 2D multi-modal features in visual encoding alongside point-level training. Consequently, we propose a dual-branch point cloud network capable of extracting 3D multi-modal features.

\subsubsection{2D-Supervised 3D-Language Alignment}
\label{sec: a}
Recent studies utilize the foundational PointNet++ \cite{qi2017pointnet++} network to process point clouds. To ensure fairness, our network similarly adopts the multilayer structure of PointNet++. This structure processes a point cloud $\mathcal{P} = \{\mathbf{p}_i\}_{i=1}^N \in \mathbb{R}^{N \times 6}$, containing XYZ coordinates and RGB colors, via hierarchical Set Abstraction (SA) layers: $\mathcal{P}^{(l+1)} = {SA}(\mathcal{P}^{(l)})$. In each layer, a specific number of seeds is determined and gathered around the sampled points using Farthest Point Sampling. Subsequently, multi-scale features are extracted through a Multilayer Perceptron (MLP). The final layer produces a unique global point cloud feature that effectively represents the object.

Most prior work encodes the semantic constraint by either adding a classification layer after the point cloud network to enhance the entity-level semantic capturing ability \cite{roh2022languagerefer, huang2022multi, guo2023viewrefer}, or by directly using closed-set instance segmentation categories as prior knowledge \cite{zhao20213dvg, yuan2021instancerefer, chang2024mikasa}. Such paradigms exhibit limited discriminative abilities for only predefined semantic classes, heavily reliant on training environments. More crucially, these methods cannot fundamentally establish correspondence between learnable 3D structural representations and open-vocabulary textual descriptions.

Our goal is to align object-level 3D vision with language using an open vocabulary. This is accomplished by integrating auxiliary supervision from large-scale 2D multi-modal models during training. Specifically, we leverage the pre-trained CLIP \cite{radford2021learning} to jointly encode visual representations and textual descriptions. The image branch of CLIP extracts 2D visual features from object-bounding boxes projected onto randomly sampled video frames, employing a single randomly selected feature per object for auxiliary training. The text branch, on the other hand, receives class-aware prompts formulated using a template: ``The object is [\textit{instance name}]'', from which textual features are derived. These instance names correspond to the object types found in the training set utterances. Following this, the 3D point cloud features are aligned with textual modalities by optimizing a contrastive loss. This loss aims to preserve both the cross-modal similarity and the inherent semantic correspondence present in CLIP. Taking inspiration from CLIP-KD \cite{yang2024clip}, the visual and textual features from the pre-trained model are further distilled into the 3D vision and text understanding tasks through interactive contrastive learning. The 3D object-text alignment loss $\mathcal{L}_{ot}$ with 2D assistant training is defined as:
\begin{align}
\mathcal{L}_{ot} = &\,
    \mathcal{L}_c(F_P, F_T)+\mathcal{L}_c(F_P, F_{I^*}) 
    + \mathcal{L}_c(F_T, F_{T^*}) \nonumber \\
    &+ \mathcal{L}_c(F_P, F_{T^*}) 
    + \mathcal{L}_c(F_{I^*}, F_T)
\end{align}
\begin{align}
\mathcal{L}_{c}(x,y)=\frac{1}{C_1} \sum_{i=1}^{C_1} -\log \frac{exp(x_i \cdot y_i)}{\sum_{j=1}^{C_2} exp(x_i\cdot y_j)}
\end{align}

where $F_P$ represents the object features from the point network, $F_T$ represents the class-aware text features from the learnable BERT-based \cite{kenton2019bert} encoder, $F_{I^*}$ and $F_{T^*}$ respectively denote 2D-object and textual features from the frozen CLIP visual encoder and CLIP text encoder. ${L}_{c}$ represents the contrastive loss function for any two feature vectors of sizes $C_1$ and $C_2$. 

\subsubsection{2D-Enhanced Feature Extraction}
\label{sec: b}
Sparsely sampled point clouds often lack the ability to represent the color and texture attributes of object surfaces adequately. To address this limitation, many studies extract 2D-view features to achieve a richer representation \cite{chen2020scanrefer, xu2024multi, zhang2024vision}. However, the process of point-level feature fusion necessitates pixel-level encoding and decoding on the projection of complete views, resulting in substantial computational costs. In contrast, our approach solely extracts object features from the most representative view and integrates them into the geometric encoding of the point cloud network. Distinct from conventional separate encoding schemes, our method seamlessly incorporates 2D features as enhancement within the intermediate layers of the point cloud network. The cross-modal feature fusion process is formulated as:
\begin{equation}
    F_{\text{att}} = F_{i\text{-layer}}  + \text{Proj}_{d}\left[ F_{I^*} \right]
    \label{eq:cross_modal_fusion}
\end{equation}
where $F_{i\text{-layer}} $ denotes the feature matrix output from the $i$-th layer of the point network, $F_{I^*}$ contains the visual feature embeddings extracted from the CLIP vision encoder. $\text{Proj}_{d}$ represents the learned linear projection operator that reduces the dimensionality of $F_{I^*}$ to $d$ channels. The fusion module exploits the complementary properties of visual semantics from the pre-trained model and geometric structures from the point network. 

Our dual-branch point cloud encoding network ultimately outputs geometrically encoded point cloud features $F_P$ and 2D-enhanced object features $F_M$. $F_P$ is trained with extra alignment loss and added to $F_M$ as object features $F_O$. Notably, our framework exclusively processes the single view with the maximal projection coverage for each object, where the projection coordinates are directly derived from the preliminary segmentation outputs. This view selection mechanism, coupled with the direct utilization of pre-segmented spatial mappings, substantially reduces computational overhead by eliminating redundant viewpoint sampling and geometric reconstruction steps.

\begin{figure*}[!htb]
    \centering{\includegraphics[width=0.7\linewidth]{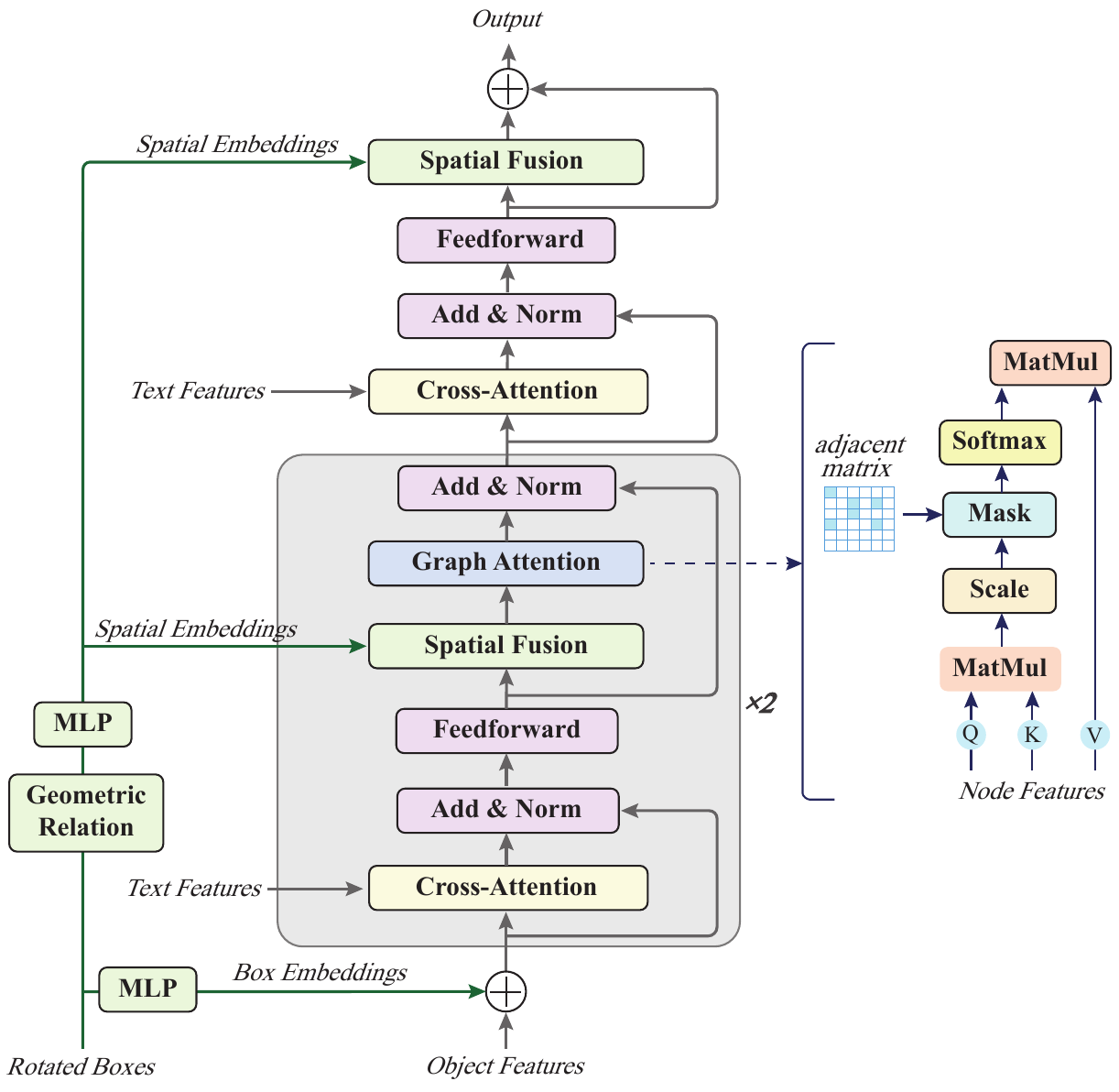}}
	\caption{\textbf{The network structure of the cross-modal interaction module.} The network mainly consists of cross-attention layers, graph attention layers, and geometry-related fusion layers. The content inside the gray box is iterated twice.}
   \vspace{-0.4cm}
    \label{fig2}
\end{figure*}
\subsection{Language-Guided Scene Graph}
\label{sec: c}
In the 3D visual grounding task, accurately distinguishing between identical objects mainly depends on their positional context and referential relationships, as described in the given description. To tackle this issue, we introduce relational graph learning. The scene graph focuses only on objects semantically related to the referential description. This reduces noise and improves reasoning efficiency. To enhance spatial reasoning capacities, we integrate object-level spatial priors with global spatial encoding \cite{chang2024mikasa}, incorporating them into the object features prior to graph attention operations. We also employ the rotated bounding box encoding paradigm \cite{huang2022multi} to facilitate multi-view spatial comprehension via decoupled shared point cloud features. 

\subsubsection{Graph Generation via Semantic Consistency}
In complex scenes, the target description often entails multiple referential relationships. The innate capability of scene graphs to capture dependencies among various nodes aids in effectively processing intricate scene compositions. Prior studies \cite{feng2021free, huang2021text, xiao2024secg} either construct fully-connected scene graphs or establish connections with neighboring nodes, implicitly learning the node and text embedding relations. However, many positional descriptions are not dependent on proximity, and a large number of objects can introduce uncertainties and extra cost in node aggregation. We propose an explicit construction of graph node connections using 3D-language matching about object semantics. First, we match 3D feature representations to class-aware prompts as described in Section \ref{sec: a}. Then, we utilize the SpaCy lemmatization module to standardize the description text by converting words to their base forms, thereby significantly improving terminological consistency across different morphological variants. Finally, we use lexemes from the same semantic category in the modified description text to identify related objects. These matched objects, considered potential targets or referred objects in localization cues, are then connected within the graph. It ensures the graph is not a generic scene representation, but a task-specific reasoning graph conditioned on the input language.

\subsubsection{Relational Learning via Graph Attention}
The graph nodes are input into a graph attention layer to facilitate multi-relational interactions, where feature representations are enhanced by aggregating heterogeneous relation information from neighboring nodes. The $i$-th node value is updated through a multi-head attention mechanism:
\begin{equation}
v_i' =v_i+\Bigg\|_{k=1}^K( \frac{1}{\sqrt{d_k}} \sigma \sum_{j \in \mathcal{N}(i)} A_{ij} \alpha_{ij}^k W^k v_j)
\label{eq:graph_attn}
\end{equation}

where $\mathcal{N}(i)$ denotes the set of neighboring nodes of vertex $i$, $A_{ij} \in \{0,1\}$ denotes the original adjacency relation between nodes $i$ and $j$, $\alpha_{ij}^k \in (0,1]$ captures the normalized attention weight from node $j$ to $i$ through the $k$-th attention head. $W^k$ is the learnable projection matrix for the $k$-th head, $\sigma(\cdot)$ specifies the non-linear activation function.

\subsection{Cross-Modal Interaction}

Our framework implements cross-modal fusion through a heterogeneous interaction module that projects textual and visual representations into a shared latent space. As shown in Figure \Cref{fig2}, the proposed module is implemented via an alternating architecture composed of cross-modal interaction and relational graph learning. To enhance the representation of object geometric positions, we use an MLP (Multilayer Perceptron) to encode box vectors that are rotated to multiple viewpoints and then fused with the original object features as input. The interaction layers establish vision-language alignment through multi-head cross-attention mechanisms and the graph-relational layers encode structural relations via graph attention networks. Noteworthy is the iterative application of the graph attention module, executed twice, with each iteration commenced by integrating geometric relational features into the input nodes. The vision-language cross-attention operation is formulated as:
\begin{equation}
\text{CrossAtt}(Q_v, K_t, V_t) = \text{softmax}\left(\frac{Q_v K_t^T}{\sqrt{d_k}}\right)V_t
\end{equation}
where $Q_v$ denotes the query matrix projected from the visual features, $K_t$ and $V_t$ denote the key and value matrix projected from the textual features, and $d_k$ is the dimension per head.

This structure ensures that the graph nodes consistently contain linguistically informed cross-modal fusion content whenever the graph network engages in relational aggregation. The proposed relational learning mechanism is constructed based on a multi-feature alignment framework that integrates semantic, spatial, and textual dimensions, which enhances the capability of the model to prioritize task-relevant nodes during iterative optimization processes selectively.

\subsection{Training}
The training of our framework involves the following three parts: (i) a dual-branch 3D point cloud encoder, (ii) a text encoder with pre-trained BERT parameters initialized, and (iii) a cross-modal interaction module that consists of graph networks and cross-attention layers. To improve the robustness of the model, we employ a stochastic view selection strategy to the 2D-enhanced branch in the object encoder. In the training stage, inputs are sampled from different viewpoints dynamically, while one representative view which has the most projected points is set as the input during inference.
The composite loss function is formulated as a weighted combination:
\begin{align}
\mathcal{L}= \lambda_1\mathcal{L}_{ot}+\mathcal{L}_{ref}+\lambda_2\mathcal{L}_{t}+\lambda_3\mathcal{L}_{of}
\end{align}
where $\mathcal{L}_{ot}$ quantifies 3D object-text alignment using the metric defined in Section \ref{sec: a}, $\mathcal{L}_{ref}$ addresses localization accuracy through predicted target probability distributions, $\mathcal{L}_{t}$ calculates a multi-class cross-entropy loss for text-based target classification, and $\mathcal{L}_{of}$ computes object-type classification loss based on fused features from the dual-branch point cloud encoder. The weighting coefficients $\lambda_1, \lambda_2, \lambda_3 \in (0,1)$ balance task-specific objectives. 

We introduce a hybrid training paradigm to mitigate error propagation stemming from object proposals from the training-free segmentation module. During the second phase of training, we employ a stochastic fusion mechanism, wherein objects are randomly chosen from either segmentation outputs or ground-truth (GT) annotations. This approach allows the model to benefit from precise geometric projections derived from validated annotations, while concurrently enhancing its robustness against uncertainties in instance segmentation.

\section{Experiment}
\subsection{Datasets and Evaluation Metrics}
\subsubsection{Nr3D/Sr3D} 
The ReferIt3D benchmark \cite{achlioptas2020referit3d} provides two large-scale visual grounding datasets on 3D scenes from the ScanNet dataset \cite{dai2017scannet}: Nr3D and Sr3D. Nr3D comprises 41,503 natural language descriptions annotated by human participants, capturing diverse linguistic expressions of object references. In contrast, Sr3D contains 83,572 synthetically generated utterances constructed through rule-based templates encoding five fundamental spatial relationships. Both datasets control scene complexity, with each target object accompanied by up to 6 distractors of the same category within the corresponding 3D environment. The ReferIt3D benchmark predefines known object-independent point clouds and only focuses on the cross-modal comprehension of the model in the 3D scene.
\subsubsection{ScanRefer} 
The ScanRefer dataset \cite{chen2020scanrefer} contains 51,583 linguistically unconstrained descriptions annotated by human annotators, corresponding to 11,046 distinct objects within ScanNet scenes \cite{dai2017scannet}. ScanRefer prefers longer utterances with multi-sentence compositional descriptions per object compared to Nr3D. Its evaluation is based on the original point cloud, so an object detector is usually required before referential grounding in the two-stage detection-localization pipeline.

\subsubsection{Evaluation Metrics} 
Current 3D visual grounding benchmarks use specific evaluation protocols designed to address their unique technical challenges. The ReferIt3D benchmark implements distractor-aware analysis measures fine-grained discrimination capability through scene partitioning into \textit{Easy} ($\leq$2 distractors) and \textit{Hard} ($\textgreater$2 distractors) subsets. It also focuses on perspective robustness about viewpoint invariance using \textit{View-dependent} (visually distinctive objects) and \textit{View-independent} (ambiguous viewpoints) configurations. Conversely, ScanRefer emphasizes precision in geometric localization by establishing the threshold of IoU (Intersection over Union) between the predicted 3D bounding box and the GT target box. To evaluate the ability to distinguish similar distractors, ScanRefer additionally divides the \textit{Multiple} ($\geq$2 same-category instances) evaluation set.

\subsection{Implementation Details}
We use the identical 3-layer PointNet++ architecture employed by prior two-stage methods to sample 1024 points per object for encoding, yielding 768-dimensional object features. We map single-view features extracted by CLIP-VIT-B/16 to a 256-dimensional space and couple them with geometric features in the third layer of the point cloud network. All multi-head attention layers in our model are configured with 8 heads. The loss weight coefficients \(\lambda_1\) and \(\lambda_3\) are set to 0.5, while \(\lambda_2\) is set to 0.1. Our training protocol leverages a NVIDIA 3090 GPU for all models, employing the Adam optimizer with a batch size of 16. Models are trained for 100 epochs on both the Nr3D and ScanRefer datasets, and for an extended 120 epochs on Sr3D. We initiate the learning rate as \(5\times 10^{-4}\), employing a scheduled multiplicative decay factor of 0.65 at intervals of 10 epochs between the 30th and 80th training epochs.

\subsection{Comparison with Others}
\subsubsection{Nr3D/Sr3D}

\begin{table*}[!t]
\begin{center}
	\caption{Comparison with other recent methods on Nr3D and Sr3D}
    \renewcommand\arraystretch{1.2}
    \fontsize{8pt}{9pt}\selectfont
    \tabcolsep=1.5pt
    \label{tab1}
    \begin{threeparttable}
	\begin{tabular}{l|c|ccccc|ccccc}
    \Xhline{2pt}
	  \multirow{2}{*}{Methods}&\multirow{2}{*}{Year}&\multicolumn{5}{c|}{\textbf{Nr3D}} &\multicolumn{5}{c}{\textbf{Sr3D}}\\
        \cline{3-12}
		&  &Overall  &Hard &Easy   &V-dep  &V-indep &Overall    &Hard &Easy  &V-dep  &V-indep \\
       \cline{1-12}
		ReferIt3D \cite{achlioptas2020referit3d} &\textit{2020}  &35.6\% &27.9\%&43.6\%   &32.5\% &37.1\% &40.8\% &31.5\% &44.7\% & 39.2\% &40.8\% \\
	  TGNN \cite{huang2021text} &\textit{2021}  &37.3\% &30.6\%&44.2\%  &35.8\% &38.0\% &45.0\% &36.9\%&48.5\%  & 45.8\% &45.0\% \\     
        3DVG-Trans \cite{zhao20213dvg} &\textit{2021} &40.8\%  &34.8\%&48.5\%  &34.8\% &43.7\% &51.4\% &44.9\%&54.2\%  &44.6\% &51.7\% \\
        FFL-3DOG \cite{feng2021free} &\textit{2021} &41.7\% &35.0\%&48.2\%   &37.1\% &44.7\% &- &- &- &- &- \\
        LAR \cite{bakr2022look} &\textit{2022} &48.9\% &42.3\%&58.4\%   &47.4\% &52.1\% &59.4\%  &51.2\% &63.0\%&50.0\% &59.1\% \\
        SAT \cite{yang2021sat}&\textit{2021}   &49.2\%  &42.4\%&56.3\%  &46.9\% &50.4\% &57.9\% &50.0\%&61.2\%  &49.2\% &58.3\% \\   
        M3DRef-CLIP \cite{zhang2023multi3drefer} &\textit{2023} &49.4\% &43.4\%&55.6\%   &42.3\% &52.9\% &- &- &- & - &- \\
        3D-SPS \cite{luo20223d} &\textit{2022} &51.5\% &45.1\%&58.1\%   &48.0\% &53.2\% &62.6\% &65.4\%&56.2\%  & 49.2\% &63.2\% \\
        EDA \cite{wu2023eda} &\textit{2023} &52.1\% &-  &- &- &- &68.1\% &- &- &- &- \\
        BUTD-DETR \cite{jain2022bottom}&\textit{2022}   &54.6\%  &48.4\%&60.7\%  &46.0\% &58.0\% &67.0\% &63.2\%&68.6\%  &53.0\% &67.6\% \\         
        MVT \cite{huang2022multi}&\textit{2022}  &55.1\%  &49.1\%&61.3\%  &54.3\% &55.4\% &64.5\% &58.8\%&66.9\%  &58.4\% &64.7\% \\
        ViewRefer \cite{guo2023viewrefer}&\textit{2023}  &56.0\%  &49.7\%&63.0\%  &55.1\% &56.8\% &67.0\% &62.1\%&68.9\%  &52.2\% &67.7\% \\   
        MCLN \cite{qian2024multi} &\textit{2024} &59.8\% &-  &- &- &- &68.4\% &- &- &- &- \\
        MiKASA \cite{chang2024mikasa} &\textit{2024} &64.4\%  &59.4\%&69.7\%  &65.4\% &64.0\% &75.2\% &67.3\%&\textbf{78.6\%}  &\textbf{70.4\%} &75.4\% \\          
        MA2TransVG \cite{xu2024multi} &\textit{2024} &65.2\% &57.6\%&71.1\%   &62.5\% &65.4\% &73.9\% &69.3\%&76.0\%  & 64.5\% &73.8\% \\      
        \hline  							
        LSVG (ours)  &- &\textbf{67.2\%}  &\textbf{61.4\%} &\textbf{73.3\%} &\textbf{65.7\%} &\textbf{68.0\%} &\textbf{76.0\%} &\textbf{71.7\%} &77.8\% &67.2\% &\textbf{76.4\%} \\
        vs. MiKASA \cite{chang2024mikasa} & &\textbf{\color{red}{$\uparrow$2.8\%}} &\textbf{\color{red}{$\uparrow$2.0\%}}&\color{red}{$\uparrow$3.6\%}&\color{red}{$\uparrow$0.3\%}&\color{red}{$\uparrow$4.0\%}&\textbf{\color{red}{$\uparrow$0.8\%}} &\textbf{\color{red}{$\uparrow$4.4\%}}&\color{darkgreen}{$\downarrow$0.8\%}&\color{darkgreen}{$\downarrow$3.2\%}&\color{red}{$\uparrow$1.0\%}\\
        vs. MA2TransVG \cite{xu2024multi} & &\textbf{\color{red}{$\uparrow$2.0\%}} &\textbf{\color{red}{$\uparrow$3.8\%}}&\color{red}{$\uparrow$2.1\%}&\color{red}{$\uparrow$3.2\%}&\color{red}{$\uparrow$2.6\%}&\textbf{\color{red}{$\uparrow$2.1\%}} &\textbf{\color{red}{$\uparrow$2.4\%}}&\color{red}{$\uparrow$1.8\%}&\color{red}{$\uparrow$2.7\%}&\color{red}{$\uparrow$2.6\%}\\
    \Xhline{2pt}   				
	\end{tabular}
 \end{threeparttable}
\end{center}
\vspace{-0.4cm}
\end{table*}

As shown in Table \ref{tab1}, our approach exceeds the state-of-the-art method on both the Nr3D and Sr3D datasets by 2.0\% and 2.1\%, respectively. The improvements of 3.8\% and 2.4\% on hard samples indicate that our model can better learn object relationships in scenes with more distractors. Regarding view-dependent samples of Sr3D, although our model is slightly inferior to MiKASA, its view fusion requires additional feature extraction from multiple rotated point clouds through data augmentation. In contrast, our framework only needs to extract features once from the point cloud in the original view, resulting in a significant reduction in computational cost.

\subsubsection{ScanRefer} 

\begin{table}[!t]
\begin{center}
    \caption{Comparison with other recent methods on Scanrefer.}
    \renewcommand\arraystretch{1.2}
    \fontsize{8pt}{9pt}\selectfont
    \tabcolsep=2pt
    \label{tab2}
    \begin{threeparttable}
	\begin{tabular}{l|cc|cc}
    \Xhline{2pt}
	  \multirow{2}{*}{Methods}&\multicolumn{2}{c|}{2D Views}&\multirow{2}{*}{Multiple}&\multirow{2}{*}{Overall}\\
        \cline{2-3}
	  &View Num &Encoder&&\\
        \hline 
        TGNN \cite{huang2021text}  &-&-  &23.18\%  &29.70\%\\
        InstanceRefer \cite{yuan2021instancerefer}  &-&-  &24.77\%  &32.93\%\\
        MVT \cite{huang2022multi} &-&-  &25.26\%  &33.26\%\\
        ViewRefer \cite{guo2023viewrefer}  &-&- &26.50\%  &33.66\%\\
        BUTD-DETR \cite{jain2022bottom} &- &-     &32.81\%  &37.05\%\\
        EDA \cite{wu2023eda}  &- &-  &37.64\%  &42.26\%\\
        MCLN \cite{qian2024multi}  &- &-  &38.41\%  &42.64\%\\     
        \hline 
        ScanRefer \cite{chen2020scanrefer}  & all &ENet   &21.11\%  &27.40\%\\
        3DVG-Trans \cite{zhao20213dvg}  &all &ENet  &28.42\%  &34.67\% \\ 
	  3D-SPS \cite{luo20223d}  &all &ENet   &29.82\%  &36.98\%\\
        3DJCG \cite{cai20223djcg}  &all&ENet   &30.82\%  &37.33\%\\
        D3Net \cite{chen2022d} &all&ENet &30.05\% &37.87\%\\
        3D-LP \cite{jin2023context}&all &ENet  &33.41\%  &39.46\%\\
        3DVLP \cite{zhang2024vision} &all &ENet   &33.40\% &40.51\%\\        
        MA2TransVG \cite{xu2024multi}  &all&ENet   &41.4\% &45.7\%\\ 
        \hline 
        M3DRef-CLIP \cite{zhang2023multi3drefer} &3&CLIP  &36.8\% &44.7\%\\ 
        LSVG (ours)&1 &CLIP &\textbf{40.87\%} &\textbf{45.37\%}\\   														
    \Xhline{2pt}
	\end{tabular}
 \end{threeparttable}
\end{center}
\end{table}
Table \ref{tab2} shows the localization results (IoU$\geq0.5$) of advanced methods on Scanrefer, encompassing both methods utilizing 2D information and those relying solely on point clouds. The column ``2D Views'' shows the number of 2D views per object (``View Num'') and the pre-trained encoder networks for 2D image processing (``Encoder''). When ``View Num'' is ``all'', it means that all 2D information in the scene views will be aggregated.

Current methods mainly employ the pre-trained ENet segmentation model to extract pixel-level features from every 2D view, which are then projected onto 3D points. While this approach enhances 2D features, it comes with significant computational costs and redundant visual data due to the high overlap among adjacent views. In contrast, our method leverages CLIP for object-level feature fusion, requiring only one image for feature extraction per object. Compared to M3DRef-CLIP, which also uses CLIP to encode 2D objects, our LSVG surpasses it with fewer images. The 4.0\% improvement on ``Multiple'' demonstrates that our approach is more effective at understanding positional relationships within the text, thereby enabling the distinction between similar objects. 

\subsection{Visualization Analysis}
\begin{figure}[!htb]
    \centering
    \includegraphics[width=1.0\linewidth]{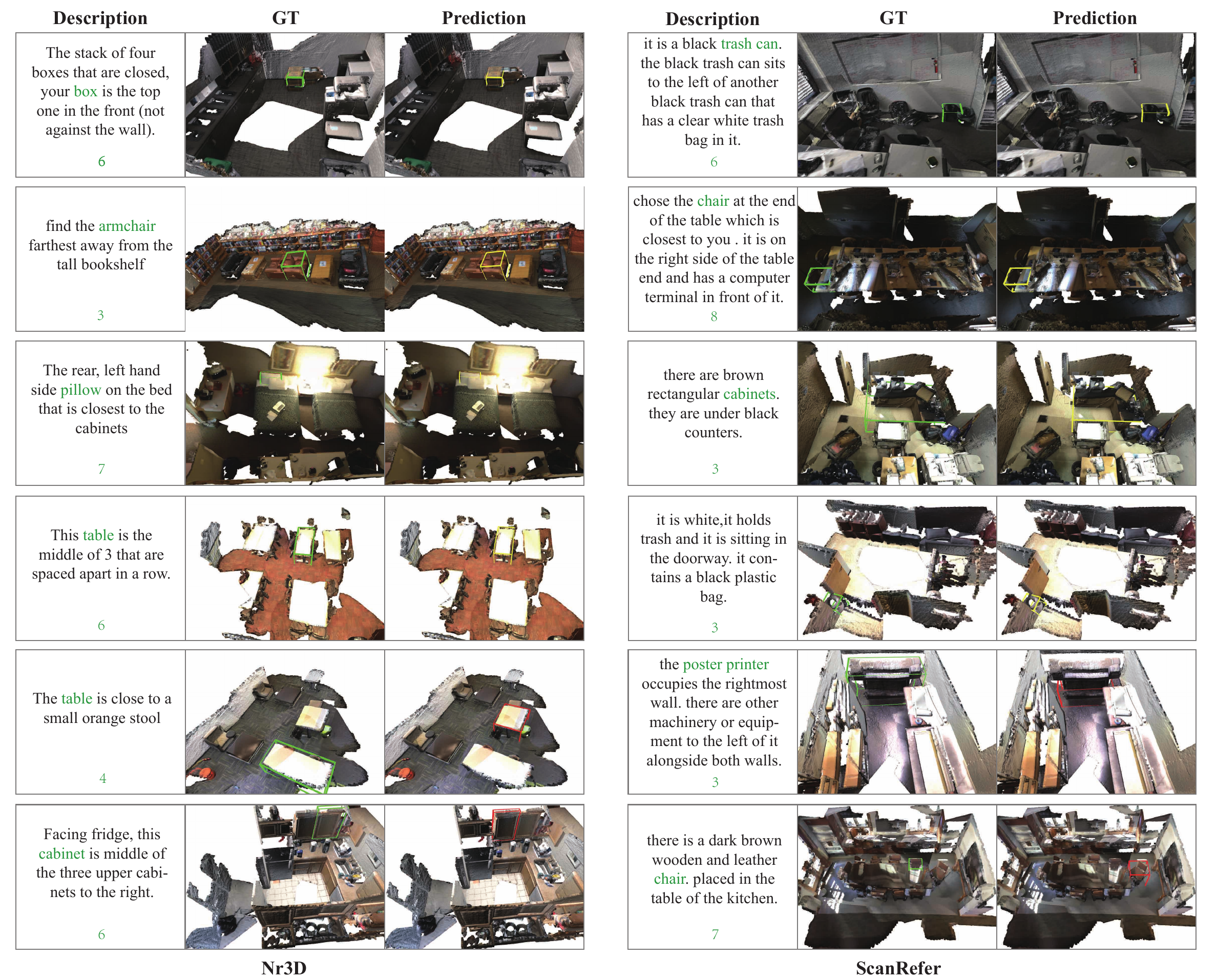}
	\caption{\textbf{Visualization of our method on Nr3D and ScanRefer.} The number under the text entry in the “Description” column indicates the number of target-class objects in the scene. Yellow bounding boxes surround the correctly predicted targets, and the error predicted results are marked with red bounding boxes.}
   \vspace{-0.4cm}
    \label{fig3}
\end{figure}

Figure \Cref{fig3} presents visualization results of our method on the Nr3D and ScanRefer datasets. Both datasets are manually annotated with free-form textual descriptions, where sentences in ScanRefer are typically longer and contain more attribute information. The localization of Nr3D is directly based on GT boxes, while ScanRefer is based on predicted boxes from the segmentation model. The results demonstrate that our proposed method achieves precise localization accuracy in challenging scenes containing over two distractors. Nevertheless, upon analyzing the failure cases, we identify two primary shortcomings of our approach: decreased sensitivity to variations in distance and a drop in performance when dealing with perspective-dependent descriptions.

\begin{table*}[!t]
\begin{center}
	\caption{Ablation studies of key components on Nr3D and ScanRefer}
    \renewcommand\arraystretch{1.2}
    \fontsize{8pt}{9pt}\selectfont
    \label{tab3}
    \begin{threeparttable}
	\begin{tabular}{c|ccc|cc}
    \Xhline{2pt}
        &\multicolumn{3}{c|}{Nr3D}&\multicolumn{2}{c}{ScanRefer}\\
        \cline{2-6} 	
	   & Overall  &Hard &V-dep  &Overall&Multiple\\
       \cline{1-6} 					
         baseline &\textbf{67.2\%}  &\textbf{61.4\%} &\textbf{65.7\%} &\textbf{45.37\%}  &\textbf{40.87\%}\\ 		
        w/o 2D supervision &65.3\% &59.5\% &64.9\%  &44.86\% &40.32\%\\ 
        w/o 2D enhancement &58.7\% &53.4\% &57.8\%  &42.50\% &37.97\%\\  
        w/o graph learning &65.0\% &58.6\% &64.3\% &44.73\% &40.10\%\\ 				
    \Xhline{2pt}  
	\end{tabular} 
 \end{threeparttable}
\end{center}
\vspace{-0.4cm}
\end{table*}

\subsection{Ablation Studies}
\subsubsection{Analysis of Key Components}
\begin{table}[!b]
\vspace{-0.4cm}
\begin{center}
	\caption{Ablation studies of 3D object-text loss on Nr3D}
    \renewcommand\arraystretch{1.4}
    \fontsize{8pt}{9pt}\selectfont
    \tabcolsep=4pt
    \label{tab4}
	\begin{tabular}{ccc|ccccc}
    \Xhline{2pt}
        $PI^*$ & $TT^*$ & $IC$ & Overall &Hard &Easy &V-dep  &V-indep\\
        \cline{1-8}   		
           \checkmark &\checkmark  &\checkmark &\textbf{67.2\%}  &\textbf{61.4\%} &\textbf{73.3\%} &\textbf{65.7\%} &\textbf{68.0\%}  \\
           & \checkmark & \checkmark   &66.2\%  & 60.3\%  &72.4\%  &64.3\%  &67.2\%\\
          \checkmark  & &\checkmark  &64.7\%& 59.2\%  &70.3\%  & 63.1\% &65.4\%   \\ 				
         \checkmark  & \checkmark & & 64.7\%  &59.0\%  &70.7\%  &64.2\%  &65.0\%\\ 		
    \Xhline{2pt}
	\end{tabular}
\end{center}
\end{table}
Table \ref{tab3} presents ablation studies for the Nr3D and ScanRefer datasets, showcasing the effectiveness of three key components: the 2D-supervised 3D-language alignment (Section \ref{sec: a}), the 2D-enhanced encoding branch (Section \ref{sec: b}), and the graph learning module based on the language-guided scene graph (Section \ref{sec: c}). The standardized bounding box annotations of Nr3D aim to separate detector performance from the evaluation metric, thereby enabling a direct assessment of referential language comprehension capabilities. The metrics of ScanRefer involve the accuracy of the 3D bounding box. Therefore, in addition to multi-modal comprehension, there is also a segmentation effect. ``w/o graph learning'' means removing all of the graph attention layers from the cross-modal interaction module. By comparing the results of ``w/o 2D supervision'' to the baseline LSVG, it is evident that 3D features trained using a 2D pre-trained multi-modal model align more easily with text. The slight improvement in viewpoint dependency challenges arises because 2D vision does not assist in perceiving viewpoint information. The results of ``w/o graph learning'' show that relational graph learning has significantly improved 3D localization based on referential descriptions, especially on hard samples with complex environments. The 2D-enhanced encoding obtained from CLIP demonstrates the most notable performance improvement across all evaluation metrics (``w/o 2D enhancement''). This underscores the effectiveness of our proposed dual-branch point cloud encoding network in significantly improving the 3D visual understanding capabilities of the multi-modal grounding model.

\begin{figure}[!htb]
    \centering{\includegraphics[width=1.0\linewidth]{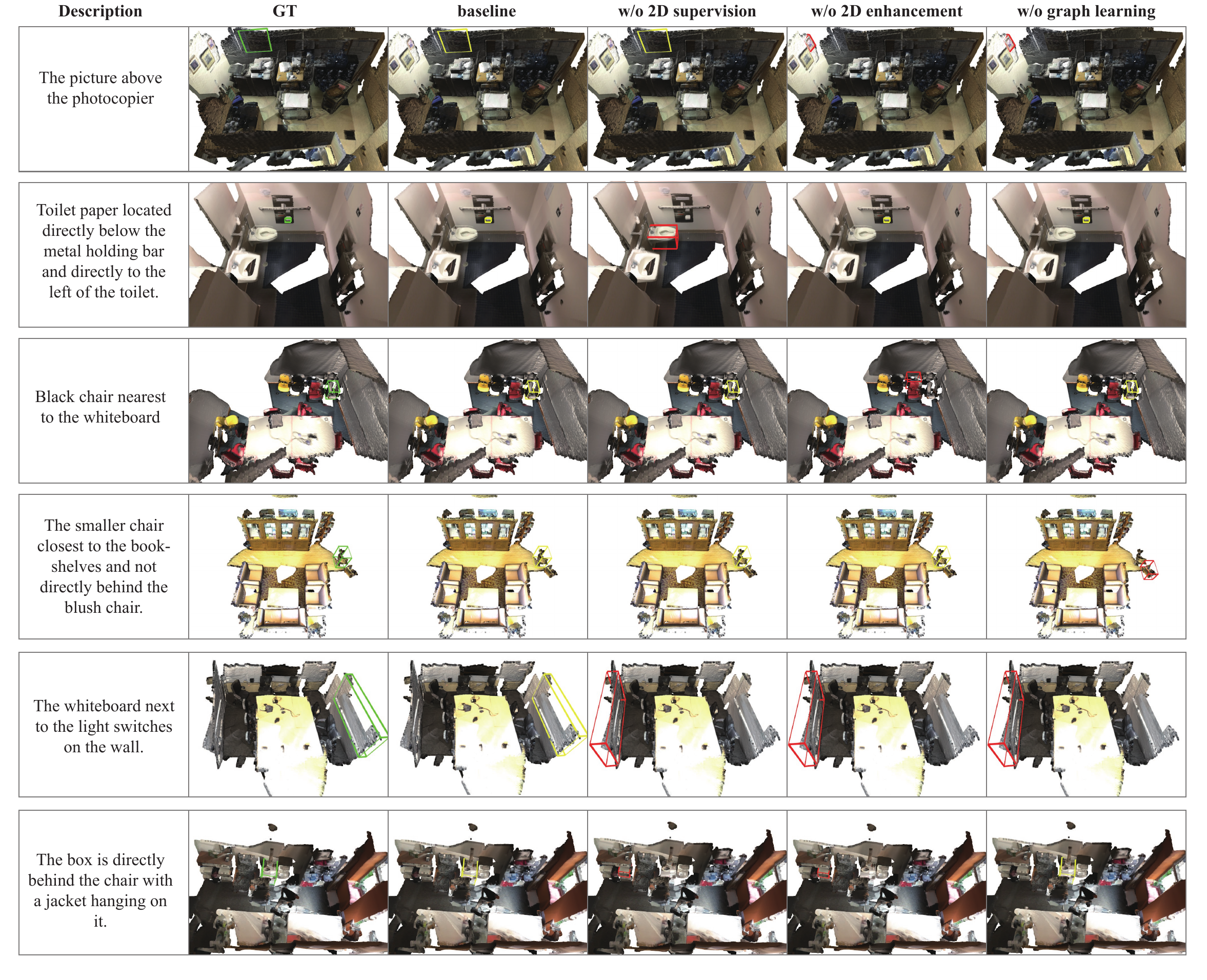}}
	\caption{\textbf{Visualization of the results of ablation experiments.} The ``baseline'' denotes the complete LSVG method we propose. The last three columns present the localization results of the model without the three important components. The incorrectly predicted 3D bounding boxes are highlighted in red.}
   \vspace{-0.4cm}
    \label{fig4}
\end{figure}

\begin{figure}[!htb]
    \centering{\includegraphics[width=0.7\linewidth]{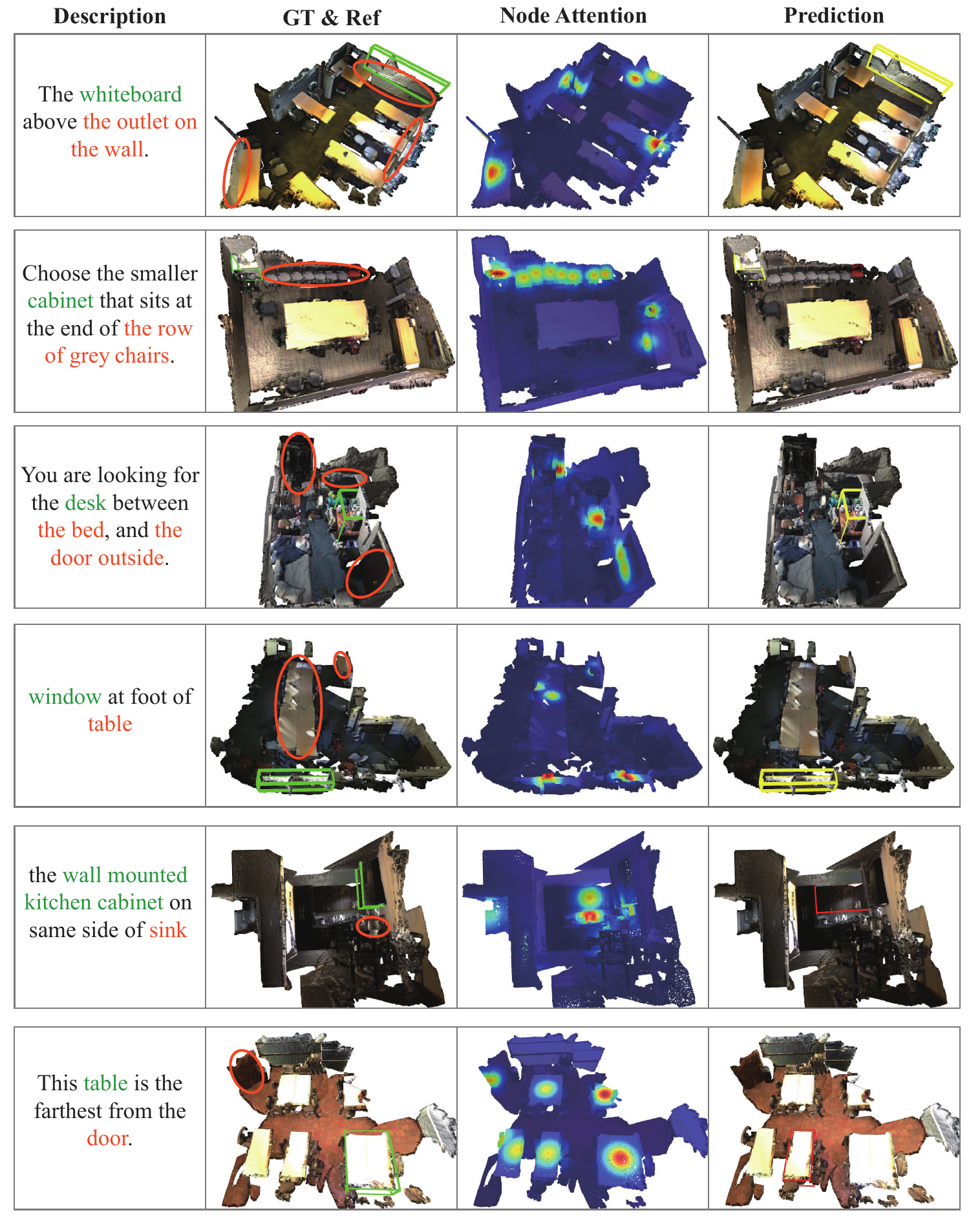}}
	\caption{\textbf{Visualization of graph attention results.} The third column shows the heat maps of the attention weights of the last graph attention layer, where only the top 10 responsive node pairs are retained to represent object scores. In the second column, Orange ellipses demarcate possible references related to a descriptive relationship. Green bounding boxes denote GT targets, while the predictions of our model are color-coded with yellow for correct and red for incorrect.}
   \vspace{-0.4cm}
    \label{fig5}
\end{figure}

The visualization results in Figure \Cref{fig4} demonstrate the role of each component in 3D visual grounding for multi-modal understanding. The 2D-assisted modules focus more on the inherent object attributes and class-aware semantics. The absence of these two components may lead to erroneous perception, such as misidentification of non-target object categories in the second-row example or ambiguity in encoding additional attributes of the referred object (``chair'') in the last-row case. Relational learning based on scene graphs primarily manifests in the comprehension of referential relationships within textual descriptions. Particularly when the target is close to homogeneous distractors (as exemplified in the fourth-row case), the graph attention mechanism demonstrates enhanced capability in distinguishing the target object through leveraging complex spatial relationships. Figure \Cref{fig5} illustrates the maximum attention scores assigned to nodes during feature aggregation in the final graph attention layer. It can be seen that for accurately predicted samples, the relational graph assigns higher attention scores to target or referred objects that match the textual description. Notably, in the last row, although the target object receives a relatively higher attention score, the aggregated features do not offer a competitive advantage in the subsequent text alignment processing, ultimately resulting in incorrect localization.

\subsubsection{Losses of 2D-Assisted Training}
In Section \ref{sec: a}, we employ various contrastive losses to define the 3D object-text alignment loss. The losses that leverage 2D pre-trained multi-modal features play an important role in 3D vision-language learning. Table \ref{tab4} presents the quantitative results of the impact of various loss functions on localization performance. ``$PI^*$'' denotes the contrastive loss between the point cloud features and CLIP-extracted 2D features. ``$TT^*$'' represents the contrastive loss between text features extracted by our framework and CLIP-generated textual embeddings. ``$IC$'' refers to the interactive contrastive loss that aligns 3D and 2D visual features with the interchangeable corresponding text features. The tabular data demonstrates that both text contrastive and interaction contrastive losses substantially improve 3D multi-modal learning. This suggests that incorporating 2D-assisted training can help mitigate the issue of cross-modality understanding in situations where 3D data is limited.

\subsubsection{Structure of Object Encoder}
\begin{table}[!t]
\begin{center}
	\caption{Ablation studies of object encoder branch structure on Nr3D }
    \renewcommand\arraystretch{1.4}
    \fontsize{8pt}{9pt}\selectfont
    \tabcolsep=4pt
    \label{tab5}
	\begin{tabular}{cc|ccccc}
    \Xhline{2pt}
         layer-$i$&att-dim & Overall &Hard &Easy &V-dep  &V-indep\\
        \cline{1-7}   		
           1& 3 & 61.0\%  &55.5\%  &66.7\%  &61.1\%  &61.0\%  \\ 				
           2& 128 & 65.7\%  & 59.9\%  &71.7\%  &64.3\%  &66.4\%\\  				
           \rowcolor{gray!20} 3& 256 &67.2\%  &61.4\% &73.3\% &65.7\% &68.0\%   \\ 	
           4& 768 & 66.8\%  &60.8\%  &73.1\%  &65.5\%  &67.5\%\\  				
    \Xhline{2pt}
	\end{tabular}
\end{center}
\vspace{-0.4cm}
\end{table}
The object encoder based on the PointNet++ network is a layer-by-layer process of geometric sampling and feature aggregation, so at which layer is it optimal to fuse 2D pre-trained features and generate the new encoding branch? Table \ref{tab5} illustrates the final localization effect on the set of 2D-enhanced encoding branches across different layers. ``layer-$i$'' denotes that the network architecture bifurcates into two parallel branches at the $i$-th layer, each dedicated to processing distinct feature representations. ``att-dim‘’ denotes the mapping dimension of the 2D object features extracted from CLIP before fusion with the point cloud features, to match the output dimension of each layer of the point cloud network. The experiment results demonstrate that the optimal fusion depth for pre-trained feature integration occurs at the third layer. At this specific layer, seed points exhibit partial dispersion within the geometric space, enabling the new branch to achieve better multi-modal representations in assimilating rich 2D contextual information into the 3D encoding framework.

\subsubsection{Number of Graph Attention Layers}

\begin{table}[!t]
\begin{center}
	\caption{Ablation studies of graph attention layers on Nr3D}
    \renewcommand\arraystretch{1.4}
    \fontsize{8pt}{9pt}\selectfont
    \tabcolsep=4pt
    \label{tab6}
	\begin{tabular}{c|ccccc}
    \Xhline{2pt}
         layer number & Overall &Hard &Easy &V-dep  &V-indep\\
        \cline{1-6}   		
           0 & 65.0\% &58.6\% &71.7\% &64.3\% &65.4\%  \\ 				
           1& 65.0\%  &59.3\% &70.9\% &64.1\% &65.4\%\\  								
           \rowcolor{gray!20} 2&67.2\%  &61.4\% &73.3\% &65.7\% &68.0\%  \\ 				
           3 & 65.4\%  &59.5\% &71.5\% &64.4\% &65.9\%\\  			 					
    \Xhline{2pt}
	\end{tabular}
\end{center}
\vspace{-0.4cm}
\end{table}
We construct a scene graph and utilize two graph attention layers within the cross-modal interaction module to aggregate node features. The learning objective of the graph network is to capture content related to the referential relationships in textual descriptions by updating node information. However, excessive feature aggregation can result in over-smoothing, which may lead to partial information loss in node representations. Table \ref{tab6} illustrates the effect of varying the number of graph attention layers on overall performance. The results show that better localization outcomes are achieved when the number of layers is two.

\subsubsection{CLIP with Different Scales}
\begin{table}[!t]
\begin{center}
	\caption{Ablation studies of various CLIP models on Nr3D }
    \renewcommand\arraystretch{1.4}
    \fontsize{8pt}{9pt}\selectfont
    \tabcolsep=4pt
    \label{tab7}
	\begin{tabular}{cc|ccccc}
    \Xhline{2pt}
         CLIP &dim & Overall &Hard &Easy &V-dep  &V-indep\\
        \cline{1-7}   		
          ViT-B/16&512  &\textbf{67.2\%}  &\textbf{61.4\%} &\textbf{73.3\%} &65.7\% &\textbf{68.0\%} \\
          ViT-B/32&512  & 65.1\%  &59.6\%  &70.8\%  &64.4\%  &65.4\%  \\ 	
          ViT-L/14&768  & 66.3\%  &60.5\%  &72.3\%  &64.1\%  &67.4\%  \\ 
          ViT-H/14&1024  & 66.8\%  &61.1\%  &72.7\%  &\textbf{66.1}\%  &67.1\%  \\  			
    \Xhline{2pt}
	\end{tabular}
\end{center}
\vspace{-0.6cm}
\end{table}

We construct experiments in Table \ref{tab7} to discuss the effect of different pre-trained CLIP models. All of the models are trained on the large-scale dataset LAION-2B. The network structure of ViT-B/16 is similar to that of ViT-B/32, and the output feature dimensions are both 512, but the image chunks are smaller ($16\times16$). ViT-L/14 and ViT-H/14 are larger-scale models that output feature vectors of 768 and 1024 dimensions, respectively. The experimental results indicate that scaling up pre-trained models to larger sizes yields no statistically significant improvement in overall performance, suggesting that compact models are sufficient for object-level training requirements. Furthermore, the ViT-B/16 architecture with finer granularity attains a better multi-modal representation compared to the ViT-B/32.

\section{Conclusion}
In this paper, we focus on modeling the referred objects for localization based on referential descriptions in complex scenes and propose a novel framework for 3D visual grounding. The framework first segments scene point clouds into class-agnostic object proposals and then incorporates a pre-trained 2D multi-modal model to enhance 3D visual encoding, enabling the semantic alignment of object features. We construct a language-guided scene graph to facilitate relational learning, where the nodes connect only with semantically related neighbors derived from the textual description. We design an iterative cross-modal interaction module that integrates graph attention layers and vision-language cross-attention layers. Finally, the most relevant object to the description is selected as the target. Experimental results on the Referit3D and ScanRefer benchmarks show that our proposed LSVG delivers better localization performance than other advanced methods, especially demonstrating notable improvement in challenging samples with multiple distractors.

\section*{Acknowledgments}
This work presented in this paper is partially supported by grant from Fundamental Research Funds for the Central Universities(No. 2024ZYGXZR104).


\bibliographystyle{elsarticle-num} 
\bibliography{literature}

\begin{thebibliography}{10}
\expandafter\ifx\csname url\endcsname\relax
  \def\url#1{\texttt{#1}}\fi
\expandafter\ifx\csname urlprefix\endcsname\relax\def\urlprefix{URL }\fi
\expandafter\ifx\csname href\endcsname\relax
  \def\href#1#2{#2} \def\path#1{#1}\fi

\bibitem{liu2025survey}
D.~Liu, Y.~Liu, W.~Huang, W.~Hu, A survey on text-guided 3-d visual grounding: Elements, recent advances, and future directions, IEEE Transactions on Neural Networks and Learning Systems (2025).

\bibitem{achlioptas2020referit3d}
P.~Achlioptas, A.~Abdelreheem, F.~Xia, M.~Elhoseiny, L.~Guibas, Referit3d: Neural listeners for fine-grained 3d object identification in real-world scenes, in: Computer Vision--ECCV 2020: 16th European Conference, Glasgow, UK, August 23--28, 2020, Proceedings, Part I 16, Springer, 2020, pp. 422--440.

\bibitem{he2021transrefer3d}
D.~He, Y.~Zhao, J.~Luo, T.~Hui, S.~Huang, A.~Zhang, S.~Liu, Transrefer3d: Entity-and-relation aware transformer for fine-grained 3d visual grounding, in: Proceedings of the 29th ACM International Conference on Multimedia, 2021, pp. 2344--2352.

\bibitem{vaswani2017attention}
A.~Vaswani, N.~Shazeer, N.~Parmar, J.~Uszkoreit, L.~Jones, A.~N. Gomez, {\L}.~Kaiser, I.~Polosukhin, Attention is all you need, Advances in neural information processing systems 30 (2017).

\bibitem{yang2021sat}
Z.~Yang, S.~Zhang, L.~Wang, J.~Luo, Sat: 2d semantics assisted training for 3d visual grounding, in: Proceedings of the IEEE/CVF International Conference on Computer Vision, 2021, pp. 1856--1866.

\bibitem{roh2022languagerefer}
J.~Roh, K.~Desingh, A.~Farhadi, D.~Fox, Languagerefer: Spatial-language model for 3d visual grounding, in: Conference on Robot Learning, PMLR, 2022, pp. 1046--1056.

\bibitem{huang2022multi}
S.~Huang, Y.~Chen, J.~Jia, L.~Wang, Multi-view transformer for 3d visual grounding, in: Proceedings of the IEEE/CVF Conference on Computer Vision and Pattern Recognition, 2022, pp. 15524--15533.

\bibitem{guo2023viewrefer}
Z.~Guo, Y.~Tang, R.~Zhang, D.~Wang, Z.~Wang, B.~Zhao, X.~Li, Viewrefer: Grasp the multi-view knowledge for 3d visual grounding, in: Proceedings of the IEEE/CVF International Conference on Computer Vision, 2023, pp. 15372--15383.

\bibitem{chen2022language}
S.~Chen, P.-L. Guhur, M.~Tapaswi, C.~Schmid, I.~Laptev, Language conditioned spatial relation reasoning for 3d object grounding, Advances in neural information processing systems 35 (2022) 20522--20535.

\bibitem{xu2024multi}
C.~Xu, Y.~Han, R.~Xu, L.~Hui, J.~Xie, J.~Yang, Multi-attribute interactions matter for 3d visual grounding, in: Proceedings of the IEEE/CVF Conference on Computer Vision and Pattern Recognition, 2024, pp. 17253--17262.

\bibitem{chang2024mikasa}
C.-P. Chang, S.~Wang, A.~Pagani, D.~Stricker, Mikasa: Multi-key-anchor \& scene-aware transformer for 3d visual grounding, in: Proceedings of the IEEE/CVF Conference on Computer Vision and Pattern Recognition, 2024, pp. 14131--14140.

\bibitem{chang2021comprehensive}
X.~Chang, P.~Ren, P.~Xu, Z.~Li, X.~Chen, A.~Hauptmann, A comprehensive survey of scene graphs: Generation and application, IEEE Transactions on Pattern Analysis and Machine Intelligence 45~(1) (2021) 1--26.

\bibitem{zhang2026scenellm}
H.~Zhang, Z.~Li, J.~Liu, Scenellm: Implicit language reasoning in llm for dynamic scene graph generation, Pattern Recognition 170 (2026) 111992.

\bibitem{feng2021free}
M.~Feng, Z.~Li, Q.~Li, L.~Zhang, X.~Zhang, G.~Zhu, H.~Zhang, Y.~Wang, A.~Mian, Free-form description guided 3d visual graph network for object grounding in point cloud, in: Proceedings of the IEEE/CVF international conference on computer vision, 2021, pp. 3722--3731.

\bibitem{li2025r2g}
Y.~Li, Z.~Wang, W.~Liang, R2g: Reasoning to ground in 3d scenes, Pattern Recognition (2025) 111728.

\bibitem{huang2021text}
P.-H. Huang, H.-H. Lee, H.-T. Chen, T.-L. Liu, Text-guided graph neural networks for referring 3d instance segmentation, in: Proceedings of the AAAI Conference on Artificial Intelligence, Vol.~35, 2021, pp. 1610--1618.

\bibitem{yuan2021instancerefer}
Z.~Yuan, X.~Yan, Y.~Liao, R.~Zhang, S.~Wang, Z.~Li, S.~Cui, Instancerefer: Cooperative holistic understanding for visual grounding on point clouds through instance multi-level contextual referring, in: Proceedings of the IEEE/CVF International Conference on Computer Vision, 2021, pp. 1791--1800.

\bibitem{lei2023recent}
Y.~Lei, Z.~Wang, F.~Chen, G.~Wang, P.~Wang, Y.~Yang, Recent advances in multi-modal 3d scene understanding: A comprehensive survey and evaluation, arXiv preprint arXiv:2310.15676 (2023).

\bibitem{chen2020scanrefer}
D.~Z. Chen, A.~X. Chang, M.~Nie{\ss}ner, Scanrefer: 3d object localization in rgb-d scans using natural language, in: European conference on computer vision, Springer, 2020, pp. 202--221.

\bibitem{zhao20213dvg}
L.~Zhao, D.~Cai, L.~Sheng, D.~Xu, 3dvg-transformer: Relation modeling for visual grounding on point clouds, in: Proceedings of the IEEE/CVF International Conference on Computer Vision, 2021, pp. 2928--2937.

\bibitem{luo20223d}
J.~Luo, J.~Fu, X.~Kong, C.~Gao, H.~Ren, H.~Shen, H.~Xia, S.~Liu, 3d-sps: Single-stage 3d visual grounding via referred point progressive selection, in: Proceedings of the IEEE/CVF Conference on Computer Vision and Pattern Recognition, 2022, pp. 16454--16463.

\bibitem{cai20223djcg}
D.~Cai, L.~Zhao, J.~Zhang, L.~Sheng, D.~Xu, 3djcg: A unified framework for joint dense captioning and visual grounding on 3d point clouds, in: Proceedings of the IEEE/CVF conference on computer vision and pattern recognition, 2022, pp. 16464--16473.

\bibitem{chen2022d}
D.~Z. Chen, Q.~Wu, M.~Nie{\ss}ner, A.~X. Chang, D 3 net: A unified speaker-listener architecture for 3d dense captioning and visual grounding, in: European Conference on Computer Vision, Springer, 2022, pp. 487--505.

\bibitem{zhang2023multi3drefer}
Y.~Zhang, Z.~Gong, A.~X. Chang, Multi3drefer: Grounding text description to multiple 3d objects, in: Proceedings of the IEEE/CVF International Conference on Computer Vision, 2023, pp. 15225--15236.

\bibitem{qi2019deep}
C.~R. Qi, O.~Litany, K.~He, L.~J. Guibas, Deep hough voting for 3d object detection in point clouds, in: proceedings of the IEEE/CVF International Conference on Computer Vision, 2019, pp. 9277--9286.

\bibitem{jiang2020pointgroup}
L.~Jiang, H.~Zhao, S.~Shi, S.~Liu, C.-W. Fu, J.~Jia, Pointgroup: Dual-set point grouping for 3d instance segmentation, in: Proceedings of the IEEE/CVF conference on computer vision and Pattern recognition, 2020, pp. 4867--4876.

\bibitem{liu2021group}
Z.~Liu, Z.~Zhang, Y.~Cao, H.~Hu, X.~Tong, Group-free 3d object detection via transformers, in: Proceedings of the IEEE/CVF international conference on computer vision, 2021, pp. 2949--2958.

\bibitem{shi2024aware}
X.~Shi, Z.~Wu, S.~Lee, Aware visual grounding in 3d scenes, in: Proceedings of the IEEE/CVF Conference on Computer Vision and Pattern Recognition, 2024, pp. 14056--14065.

\bibitem{paszke2016enet}
A.~Paszke, A.~Chaurasia, S.~Kim, E.~Culurciello, Enet: A deep neural network architecture for real-time semantic segmentation, arXiv preprint arXiv:1606.02147 (2016).

\bibitem{salvador2016faster}
A.~Salvador, X.~Gir{\'o}-i Nieto, F.~Marqu{\'e}s, S.~Satoh, Faster r-cnn features for instance search, in: Proceedings of the IEEE conference on computer vision and pattern recognition workshops, 2016, pp. 9--16.

\bibitem{bakr2022look}
E.~Bakr, Y.~Alsaedy, M.~Elhoseiny, Look around and refer: 2d synthetic semantics knowledge distillation for 3d visual grounding, Advances in neural information processing systems 35 (2022) 37146--37158.

\bibitem{ke2025language}
X.~Ke, P.~Xu, W.~Guo, Language--image consistency augmentation and distillation network for visual grounding, Pattern Recognition 166 (2025) 111663.

\bibitem{chen2023vlp}
F.-L. Chen, D.-Z. Zhang, M.-L. Han, X.-Y. Chen, J.~Shi, S.~Xu, B.~Xu, Vlp: A survey on vision-language pre-training, Machine Intelligence Research 20~(1) (2023) 38--56.

\bibitem{radford2021learning}
A.~Radford, J.~W. Kim, C.~Hallacy, A.~Ramesh, G.~Goh, S.~Agarwal, G.~Sastry, A.~Askell, P.~Mishkin, J.~Clark, et~al., Learning transferable visual models from natural language supervision, in: International conference on machine learning, PmLR, 2021, pp. 8748--8763.

\bibitem{jia2021scaling}
C.~Jia, Y.~Yang, Y.~Xia, Y.-T. Chen, Z.~Parekh, H.~Pham, Q.~Le, Y.-H. Sung, Z.~Li, T.~Duerig, Scaling up visual and vision-language representation learning with noisy text supervision, in: International conference on machine learning, PMLR, 2021, pp. 4904--4916.

\bibitem{li2021align}
J.~Li, R.~Selvaraju, A.~Gotmare, S.~Joty, C.~Xiong, S.~C.~H. Hoi, Align before fuse: Vision and language representation learning with momentum distillation, Advances in neural information processing systems 34 (2021) 9694--9705.

\bibitem{chen2023clip2scene}
R.~Chen, Y.~Liu, L.~Kong, X.~Zhu, Y.~Ma, Y.~Li, Y.~Hou, Y.~Qiao, W.~Wang, Clip2scene: Towards label-efficient 3d scene understanding by clip, in: Proceedings of the IEEE/CVF Conference on Computer Vision and Pattern Recognition, 2023, pp. 7020--7030.

\bibitem{jia2024sceneverse}
B.~Jia, Y.~Chen, H.~Yu, Y.~Wang, X.~Niu, T.~Liu, Q.~Li, S.~Huang, Sceneverse: Scaling 3d vision-language learning for grounded scene understanding, in: European Conference on Computer Vision, Springer, 2024, pp. 289--310.

\bibitem{xue2023ulip}
L.~Xue, M.~Gao, C.~Xing, R.~Mart{\'\i}n-Mart{\'\i}n, J.~Wu, C.~Xiong, R.~Xu, J.~C. Niebles, S.~Savarese, Ulip: Learning a unified representation of language, images, and point clouds for 3d understanding, in: Proceedings of the IEEE/CVF conference on computer vision and pattern recognition, 2023, pp. 1179--1189.

\bibitem{yan2024maskclustering}
M.~Yan, J.~Zhang, Y.~Zhu, H.~Wang, Maskclustering: View consensus based mask graph clustering for open-vocabulary 3d instance segmentation, in: Proceedings of the IEEE/CVF Conference on Computer Vision and Pattern Recognition, 2024, pp. 28274--28284.

\bibitem{qi2017pointnet++}
C.~R. Qi, L.~Yi, H.~Su, L.~J. Guibas, Pointnet++: Deep hierarchical feature learning on point sets in a metric space, Advances in neural information processing systems 30 (2017).

\bibitem{yang2024clip}
C.~Yang, Z.~An, L.~Huang, J.~Bi, X.~Yu, H.~Yang, B.~Diao, Y.~Xu, Clip-kd: An empirical study of clip model distillation, in: Proceedings of the IEEE/CVF Conference on Computer Vision and Pattern Recognition, 2024, pp. 15952--15962.

\bibitem{kenton2019bert}
J.~D. M.-W.~C. Kenton, L.~K. Toutanova, Bert: Pre-training of deep bidirectional transformers for language understanding, in: Proceedings of naacL-HLT, Vol.~1, 2019, p.~2.

\bibitem{zhang2024vision}
T.~Zhang, S.~He, T.~Dai, Z.~Wang, B.~Chen, S.-T. Xia, Vision-language pre-training with object contrastive learning for 3d scene understanding, in: Proceedings of the AAAI Conference on Artificial Intelligence, Vol.~38, 2024, pp. 7296--7304.

\bibitem{xiao2024secg}
F.~Xiao, H.~Xu, Q.~Wu, W.~Kang, Secg: Semantic-enhanced 3d visual grounding via cross-modal graph attention, arXiv preprint arXiv:2403.08182 (2024).

\bibitem{dai2017scannet}
A.~Dai, A.~X. Chang, M.~Savva, M.~Halber, T.~Funkhouser, M.~Nie{\ss}ner, Scannet: Richly-annotated 3d reconstructions of indoor scenes, in: Proceedings of the IEEE conference on computer vision and pattern recognition, 2017, pp. 5828--5839.

\bibitem{wu2023eda}
Y.~Wu, X.~Cheng, R.~Zhang, Z.~Cheng, J.~Zhang, Eda: Explicit text-decoupling and dense alignment for 3d visual grounding, in: Proceedings of the IEEE/CVF Conference on Computer Vision and Pattern Recognition, 2023, pp. 19231--19242.

\bibitem{jain2022bottom}
A.~Jain, N.~Gkanatsios, I.~Mediratta, K.~Fragkiadaki, Bottom up top down detection transformers for language grounding in images and point clouds, in: European Conference on Computer Vision, Springer, 2022, pp. 417--433.

\bibitem{qian2024multi}
Z.~Qian, Y.~Ma, Z.~Lin, J.~Ji, X.~Zheng, X.~Sun, R.~Ji, Multi-branch collaborative learning network for 3d visual grounding, in: European Conference on Computer Vision, Springer, 2024, pp. 381--398.

\bibitem{jin2023context}
Z.~Jin, M.~Hayat, Y.~Yang, Y.~Guo, Y.~Lei, Context-aware alignment and mutual masking for 3d-language pre-training, in: Proceedings of the IEEE/CVF Conference on Computer Vision and Pattern Recognition, 2023, pp. 10984--10994.

\end{thebibliography}



\end{document}